\documentclass{article}

     \PassOptionsToPackage{numbers, compress}{natbib}

\usepackage[preprint,nonatbib]{neurips_2023}

\usepackage[utf8]{inputenc} 
\usepackage[T1]{fontenc}    
\usepackage{hyperref}       
\usepackage{url}            
\usepackage{booktabs}       
\usepackage{amsfonts}       
\usepackage{nicefrac}       
\usepackage{microtype}      
\usepackage{xcolor}         

\usepackage{algorithm} 
\usepackage{algorithmicx}
\usepackage{algpseudocode}

\usepackage{hyperref}
\usepackage{url}
\usepackage[utf8]{inputenc} 
\usepackage[T1]{fontenc}    
\usepackage{nicefrac}       
\usepackage{multirow}
\usepackage{subcaption}
\usepackage{xspace}
\usepackage{amsmath}
\usepackage{svg}
\usepackage{bbm}
\usepackage{dsfont}

\newcommand{\ie}{\textit{i.e.\xspace}}
\newcommand{\eg}{\textit{e.g.\xspace}}
\newcommand{\etal}{\textit{et al.\xspace}}

\title{Neuro-Inspired Fragmentation and Recall to Overcome Catastrophic Forgetting in Curiosity}

\author{%
 Jaedong Hwang \quad\quad Zhang-Wei Hong \quad\quad Eric Chen\\
 \textbf{Akhilan Boopathy \quad\quad Pulkit Agrawal \quad\quad Ila Fiete }\\
Massachusetts Institute of Technology \\
  \texttt{\{jdhwang, zwhong, ericrc, akhilan, pulkitag, fiete\}@mit.edu}  \\
}

\begin{document}

\maketitle

\begin{abstract}

Deep reinforcement learning methods exhibit impressive performance on a range of tasks, but still struggle on hard exploration tasks in large environments with sparse rewards. To address this, intrinsic rewards can be generated using forward model prediction errors that decrease as the environment becomes known, and incentivize an agent to explore novel states. While prediction-based intrinsic rewards can help agents solve hard exploration tasks, they can suffer from catastrophic forgetting and actually increase at visited states.  
We first examine the conditions and causes of catastrophic forgetting in grid world environments.
We then propose a new method \emph{FARCuriosity}, inspired by how humans and animals learn. The method depends on fragmentation and recall:
an agent fragments an environment based on surprisal, and uses different local curiosity modules~(prediction-based intrinsic reward functions) for each fragment so that modules are not trained on the entire environment. 
At each fragmentation event, the agent stores the current module in long-term memory (LTM) and either initializes a new module or recalls a previously stored module based on its match with the current state. 
With fragmentation and recall, FARCuriosity achieves less forgetting and better overall performance in games with varied and heterogeneous environments in the Atari benchmark suite of tasks.
Thus, this work highlights the problem of catastrophic forgetting in prediction-based curiosity methods and proposes a solution\footnote{\url{https://github.com/FieteLab/FARCuriosity}}.

\end{abstract}


\section{Introduction}\label{sec:intro}

Reinforcement learning (RL) methods aim to learn a policy that maximizes the sum of rewards received by the agent.
The policy function, which determines the agent's actions, is usually trained to maximize a reward collected from the environment (extrinsic reward).
At each time step, the agent must decide whether to exploit known strategies that guarantee certain rewards or explore to find a better strategy that could potentially yield higher rewards than the known strategies. 
Pure exploitation can result in the agent being trapped in local optima, while excessive exploration can prohibit the agent from maximizing rewards.
This trade-off is called the \textit{exploration-exploitation} dilemma.

Many real-world tasks do not supply extrinsic rewards at every time-step, making it difficult for the agent to assess the effectiveness of different actions: rewards might only be available after a long sequence of observations and actions. This \textit{sparse reward} setting makes learning a policy challenging, and it can be helpful to add a dense and internally generated auxiliary reward (called an \textit{intrinsic reward}) that promotes exploration. Many works demonstrate the utility of intrinsic rewards, and there are a variety of popular methods for generating such rewards ~\cite{bellemare2016unifying,burda2019exploration,pathak2017curiosity}.
Most count-based intrinsic reward functions~\cite{achiam2017surprise, bellemare2016unifying,ecoffet2021first,strehl2008analysis,tang2017exploration} define the intrinsic reward of a state proportional to the inverse square of the number of visits to that state. Other models use prediction errors as a measure of novelty and thus intrinsic reward: Intrinsic Curiosity Module~(ICM)~\cite{pathak2017curiosity} employ a forward and an inverse model to predict future states and action consequences, and high intrinsic rewards are given to transitions that result in high prediction error~\cite{schmidhuber1991possibility}.
As the forward model improves its predictions, the intrinsic reward decreases. RND (Random Network Distillation)~\cite{burda2019exploration} learns to predict the next state via a random feature space.
Count-and prediction-also based models have been combined~\cite{raileanu2020ride}, and goal-based methods have been recently studied~\cite{campero2021learning}.

\begin{figure}[t!]
\begin{center}
\begin{subfigure}{0.25\linewidth}
    \centering
    \includegraphics[width=0.75\linewidth]{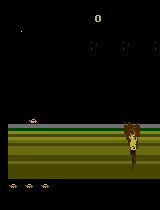}
    \caption{Starting observation} \label{fig:env_reset}
  \end{subfigure}
  \hspace{0.8cm}
    \begin{subfigure}{0.44\linewidth}
    \includegraphics[width=\linewidth]{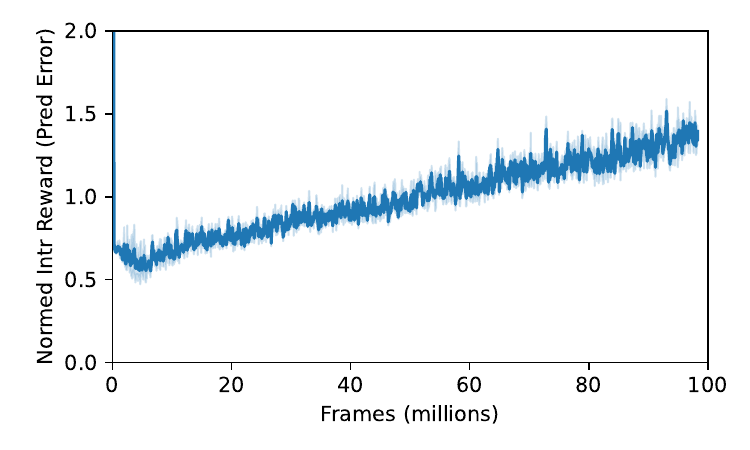}
    \caption{Intrinsic reward of (a)} \label{fig:curve}
  \end{subfigure}
\end{center}
   \caption{
   Intrinsic reward generated by RND~\cite{burda2019exploration}  during training at the first observation in each episode of Jamesbond, which is the same across episodes.
   Rewards are normalized separately using the average intrinsic reward across training, in order to directly compare the magnitude of the change in intrinsic reward. Intrinsic reward from RND continues to increase at the start location across training and repeated observations, rather than decreasing. 
   }
\label{fig:catastrophic_forgetting}
\end{figure}

Models that depend on learning to generate intrinsic reward (e.g. prediction-based models including RND and ICM) rely on the model not forgetting information about previously visited states (catastrophic forgetting ~\cite{french1999catastrophic,robins1995catastrophic}).
The issue of forgetting in prediction-based intrinsic reward models is usually overlooked, possibly because of the standard reinforcement learning paradigm in which each episode is reset to the same starting observation. 
However, as we show in Figure~\ref{fig:catastrophic_forgetting}, the intrinsic reward for the same starting observation on each episode computed by RND throughout training on Jamesbond continues to increase. This suggests that the starting observation is gradually forgotten over the course of learning within the episode.

We analyze when and why catastrophic forgetting happens even in the standard RL paradigm with environment resets at each episode, and suggest one possible relaxation inspired by how natural agents store information.
The episodic memory of natural agents divides their continuous experience of the world into multiple episodes (fragments) based on large changes in context~\cite{baldassano2017discovering,ezzyat2011constitutes,newtson1976perceptual,richmond2017constructing,swallow2009event, zacks2007event}.
It plays an important role in learning and cognition~\cite{de1946het,egan1979chunking,gobet2001chunking,gobet1998expert,simon1974big}.
Hence, we propose a fragmentation-and-recall-based curiosity (\emph{FARCuriosity}) to tackle catastrophic forgetting in the prediction-based intrinsic reward function.
While an agent explores an environment, a curiosity module generates an intrinsic reward (prediction error).
Based on this, the agent decides to store the current module in long-term memory (LTM) and initialize a new one (\emph{fragmentation}).
Alternatively, the agent refers to LTM and \emph{recalls} an existing curiosity module if the current observation is similar to one when the fragmentation happened before.
Consequently, each curiosity module does not train from the entire space of an environment but trains from a  similar subspace, and it reduces the hazard of catastrophic forgetting.

Experimental results show that  FARCuriosity achieves better performance than the baseline fragmentation-less curiosity module~\cite{burda2019exploration} on heterogeneous environments that have diverse states, leading huge distribution shifts among standard Atari game benchmarks~\cite{bellemare2013ale}.

The contribution of this paper is three-fold as follows:
\begin{itemize}
    \item We first find a catastrophic forgetting issue in the prediction-based intrinsic reward function that was overlooked due to the reset property of reinforcement learning environments.
	\item We propose a new off-the-shelf method for exploration based on \emph{fragmentation-and-recall} that divides the exploration space into multiple fragments and recalls previously explored ones.
	\item The proposed method~(\emph{FARCuriosity}) avoids catastrophic forgetting and achieves better performance compared to the baseline in heterogeneous environments.
\end{itemize}

\section{Related Work}\label{sec:related}

\subsection{Curiosity-Driven Reinforcement Learning}\label{subsec:related_rl}
Motivated by the insufficiency of rewards alone to guide exploration in sparsely rewarded  RL environments, several works construct \emph{intrinsic rewards} encouraging agents to explore unfamiliar parts of an environment.
Count-based methods provide intrinsic rewards for reaching infrequently visited states.
These methods have strong theoretical guarantees for multi-arm bandits~\cite{auer2003using} and can be effective for reinforcement learning~\cite{bellemare2016unifying,ecoffet2021first,tang2017exploration,strehl2008analysis}.
Prediction-based methods estimate where the agent cannot predict some aspect of the environment, and provide intrinsic rewards for visiting those states~\cite{achiam2017surprise,burda2019exploration,choshen2018dora,houthooft2017vime,pathak2017curiosity,raileanu2020ride}.
ICM~\cite{pathak2017curiosity}, for instance, learns a forward model to predict the next state given the current state-action pair and provides rewards scaling with the state's prediction error.
On the other hand, Intelligent Adaptive Curiosity~(IAC)~\cite{oudeyer2007intrinsic} divides sensorimotor space (state and action pairs) into multiple regions and allocates each expert module to the regions.
Although IAC requires memorizing all exemplars and directly using the similarity of exemplars as a fragmentation criterion, FARCuriosity does not need to memorize them and use prediction error (surprisal) as the criterion.

\subsection{Memory-Based Reinforcement Learning}
Memory-based reinforcement learning aims to solve the long-term credit assignment problem.
Hung~\etal~\cite{hung2019optimizing} combine LSTM~\cite{hochreiter1997long} with external memory along with an encoder and decoder for the memory.
HCAM~\cite{lampinen2021towards} uses a hierarchical LTM with chunks and attention for long-term recall inspired by Transformers~\cite{vaswani2017attention}; however, its chunks are formed periodically rather than based on content and are not used as intrinsic options for exploration.
Memory is also used for improving exploration: Go-Explore~\cite{ecoffet2021first} remembers promising states so that it can directly visit those states, where the definition of promising can include rewarded states.
Episodic Curiosity~\cite{savinov2018episodic} memorizes visited states for calculating reachability to penalize an action that leads to reachable future states.
While earlier memory-based exploration methods use memory as a guide for evaluating the intrinsic reward of a state, we augment existing exploration methods with memory, in combination with map or model {fragmentation}, to more efficiently explore.

\section{Catastrophic Forgetting in Prediction-Based Intrinsic Reward Function}\label{sec:catastrophic}

\subsection{Catastrophic Forgetting}\label{subsec:catastrophic}
Catastrophic forgetting denotes the phenomenon where neural networks lose previously learned information while learning new information~\cite{french1999catastrophic,robins1995catastrophic}.
This problem has surfaced as an important issue in `continual learning' task settings.
In continual learning, the network is trained sequentially on a set of tasks.
Each task is typically composed of disjoint subsets (e.g. classes) of datasets which may consist of a set of classes as in class-incremental learning~\cite{douillard2020podnet,park2021class,rebuffi2017icarl}.
On the other hand, Wołczyk~\etal~\cite{wolczyk2021continual} proposed a continual reinforcement learning (CRL) benchmark: an agent is sequentially trained on multiple robot manipulation tasks in Meta-World~\cite{yu2020meta} and catastrophic forgetting across tasks is calculated.

Catastrophic forgetting is measured as negative `backward transfer (BWT)', which quantifies the performance drop on earlier tasks after training on later tasks~\cite{lopez2017gradient}:
\begin{equation}
    \text{BWT} = \left( \frac{1}{T-1} \sum^{T-1}_{i=1} \xi_{T, i} - \xi_{i,i} \right),
\end{equation}
where $T$ is the number of tasks and $\xi_{j, i}$ denotes the performance (\eg, accuracy) of task $i$ after finishing task $j$.

\begin{figure}[t!]
\begin{center}
\begin{subfigure}{0.32\textwidth}
\centering
\includegraphics[width=\linewidth]{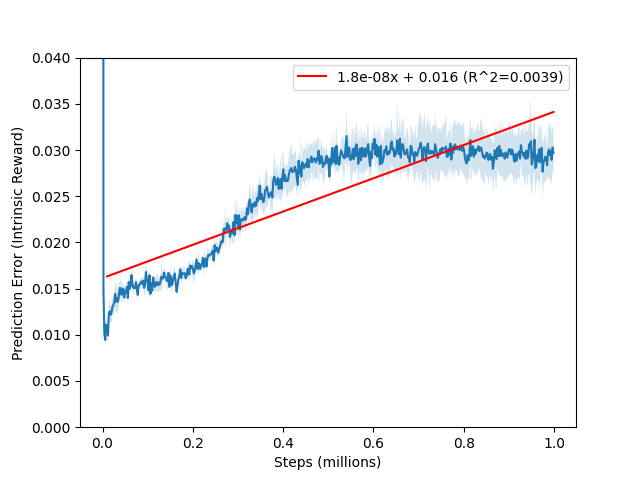}
\caption{reset-free}
\label{fig:grid_example_reset_free}
\end{subfigure}
\begin{subfigure}{0.32\textwidth}
\centering
\includegraphics[width=\linewidth]{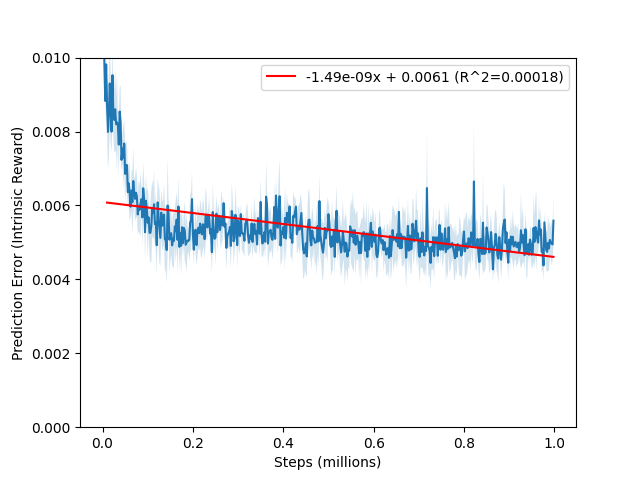}
\caption{fixed episode length}
\label{fig:grid_example_fixed}
\end{subfigure}
\begin{subfigure}{0.34\textwidth}
\centering
\includegraphics[width=0.941\linewidth]{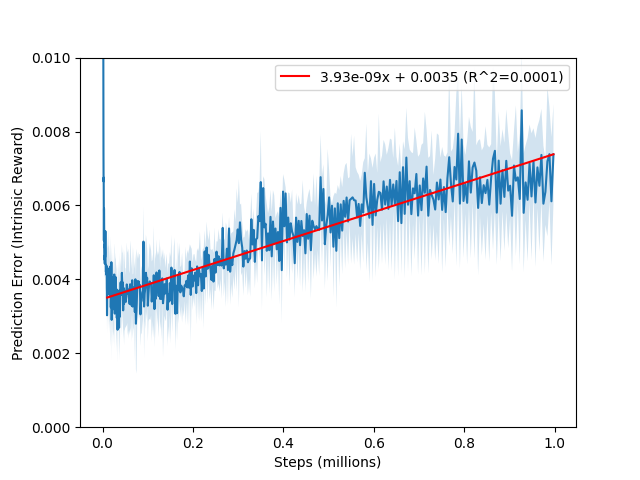}
\caption{gradually increased episode length}
\label{fig:grid_example_increased}
\end{subfigure}
\end{center}
   \caption{
   Prediction error (intrinsic reward) of the starting observation in a $10 \times 10$ grid environment with (a) reset-free, (b) fixed episode length, and (c) gradually increased episode length.
   The prediction error in reset-free environments is computed every 200 steps with the starting observation while others are measured when the agent visits there.
   The error is increased in reset-free environments and when the episode length is gradually increased, which means catastrophic forgetting happens.
}
\label{fig:grid_example}
\end{figure}

\subsection{Prediction-Based Intrinsic Reward Function}\label{subsec:catastrophic_curiosity}
In reinforcement learning, the intrinsic reward is usually designed to encourage the agent to explore novel states.
It is especially important in a sparse reward environment that the agent is hard to get any reward during training.
There is a key assumption in intrinsic reward functions:
The intrinsic reward function is a monotonically decreasing function for a given state $s$:
\begin{align}
r^i_{t_1}(s) \leq r^i_{t_2}(s) \Longleftrightarrow t_1 > t_2,
\label{eq:assumption1}
\end{align}
where $r^i_t(s)$ denotes the intrinsic reward for a state $s$ at time $t$. s
This leads the agent to explore novel (unvisited) states instead of already visited states.

In count-based exploration methods~\cite{bellemare2016unifying,strehl2008analysis,tang2017exploration},
the intrinsic reward function is defined as the inverse of the square root of the number of visits (or pseudo-count).
However, prediction-based intrinsic reward functions~\cite{burda2019exploration,pathak2017curiosity,pathak2019self} define the prediction error of the neural network for the state (or state-action pair) as an intrinsic reward.
As the training goes on, the function generates lower rewards for states that are visited multiple times.
Consequently, it seems that prediction-based intrinsic reward methods follow the assumption mentioned above.
However, this is true only if the network does not suffer from catastrophic forgetting that makes the network lose information about already visited states.
Standard reinforcement learning benchmarks~\cite{bellemare2013ale,kuettler2020nethack,lee2021ikea,szot2021habitat,tassa2018deepmind,todorov2012mujoco} usually allow resetting the environment multiple times during training so that the agent can explore different subsets of the observation space.
This seems to prevent catastrophic forgetting~(increment of intrinsic reward for the previously visited states) since the agent has a chance to visit the same state~(\eg, the starting state) during training.

Nevertheless, catastrophic forgetting can occur especially in multi-stage environments such as Jamesbond in Atari benchmark.
In the games, the agent moves in one direction and cannot go back during the episode. 
As the agent is trained, it visits more states and lives longer in each episode.
This can lead to an increase in prediction errors since the network is updated more times using other observations before observing the same state.
In other words, the non-i.i.d property of the reinforcement learning environment causes catastrophic forgetting.

\subsection{Cause of Catastrophic Forgetting}\label{subsec:toy}
We illustrate when and why catastrophic forgetting can happen with a toy experiment.
We generate a $10 \times 10$ grid environment where the observation of each location is a randomly generated 32-dimensional vector.
The agent starts from the center of the environment and randomly moves up, down, left, and right.
However, the agent is prohibited from visiting the start location.
When the environment is reset, the agent starts from the center of the environment again.
We train RND~\cite{burda2019exploration} module with random policy in this environment for one million steps sequentially.
Please refer to Section~\ref{subsec:supp_toy} for the details.

Figure~\ref{fig:grid_example} presents the intrinsic reward (prediction error) changes of the starting point across the training with various scenarios: reset-free, fixed episode length, and gradually increased episode length.
In the reset-free setting~(Figure~\ref{fig:grid_example_reset_free}), we store an observation of the starting position and keep track of the corresponding intrinsic reward during training to compare with other settings.
The intrinsic reward in this setting increases since the model is not trained on the start observation throughout training except for the first time step.
On the other hand, the reward keeps decreasing in fixed episode length (200) in Figure~\ref{fig:grid_example_fixed}, but if we increase the episode length, it will be the same as the reset-free setting.
Although the environment is reset, if the episode length is gradually increased, the intrinsic reward increases as shown in Figure~\ref{fig:grid_example_increased}.
This is because as the episode length increases, the network is updated using different observations, which leads to the network forgetting the observation in the start position.
Although this example only shows the starting observation for simplification, this scenario can be generalizable to all observations that are hard to reach during episodes.
Furthermore, in most real-world environments, it is difficult to go back to the previously visited states; although we can reset the agent, or it is impossible to reset the environment.
Hence, catastrophic forgetting is an issue that may need to be considered when training an agent using a prediction-based intrinsic reward function.

\subsection{Hazard of Catastrophic Forgetting}\label{subsec:minigrid}
To investigate the risks of catastrophic forgetting, we study the relationship between forgetting and performance drops.
To manipulate the degree of catastrophic forgetting and reduce the variance from the environment, we use count-based exploration rather than a prediction-based model in \
\textsf{MultiRoom-N6} of MiniGrid~\cite{gym_minigrid}.
The intrinsic reward of count-based exploration is defined as $1/\sqrt{N_\text{visit}(o)}$, where $N_\text{visit}(o)$ denotes the number of visits to the observation, $o$~\cite{strehl2008analysis}.
Here, we add exponential decay to the number of visits with a factor of $\gamma$ on every iteration:
\begin{equation}
    N_\text{visit}(o)  \leftarrow \gamma \cdot N_\text{visit}(o) + b,
\end{equation}
where $b$ is 1 if an agent visits $o$ at that time, otherwise 0. 
This exponential decay presents a forgetting effect that models the memory retention of short-term memory of natural agents~\cite{zhang2005visual}.
Please refer to Appendix~\ref{subsec:supp_minigrid} for experimental details.

\begin{figure}[t!]
\begin{center}
\begin{subfigure}{0.42\textwidth}
\centering
\includegraphics[width=\linewidth]{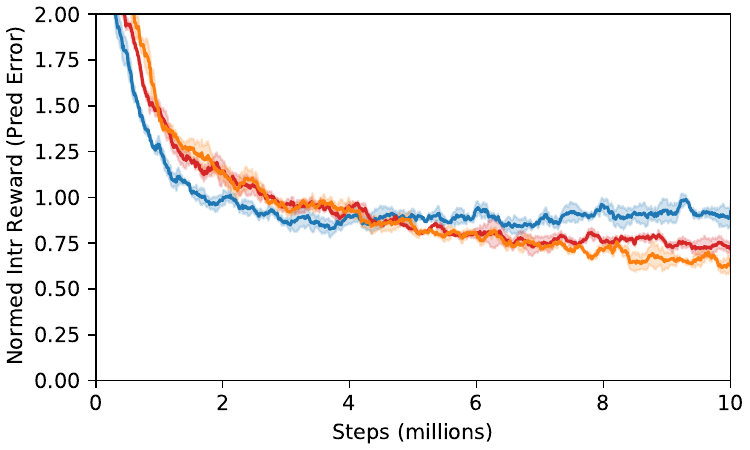}
\caption{Intrinsic reward}
\label{fig:minigrid_intr}
\end{subfigure}
\begin{subfigure}{0.42\textwidth}
\centering
\includegraphics[width=\linewidth]{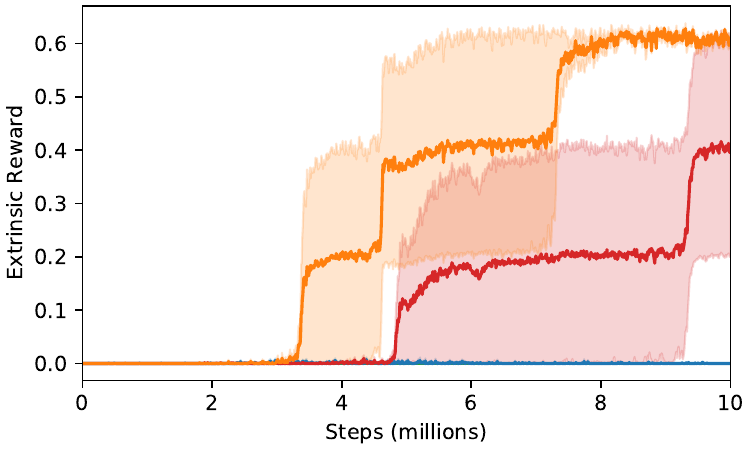}
\caption{Extrinsic reward}
\label{fig:minigrid_extr}
\end{subfigure}
\begin{subfigure}{0.14\textwidth}
\centering
\includegraphics[width=\linewidth]{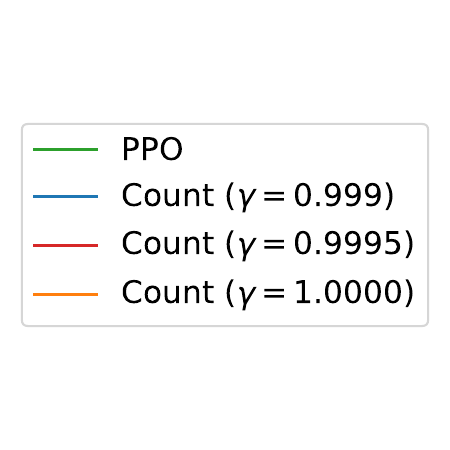}
\end{subfigure}
\end{center}
   \caption{
   (a) Intrinsic of the first observation in every episode, and (b) extrinsic reward in $\textsf{MultiRoom-N6}$ in MiniGrid benchmark during training generated by four different models: PPO and count-based exploration with various exponential decay $\gamma$.
   Even small forgetting (exponential decay) dramatically degrades the performance.
   }
\label{fig:minigrid}
\end{figure}

Figure~\ref{fig:minigrid} shows the mean cumulative extrinsic rewards of PPO with count-based exploration in \textsf{MultiRoom-N6}.
Count-based exploration without exponential decay achieves 0.6 within 10 million steps while vanilla PPO cannot learn this task.
Count-based exploration with $\gamma$ as 0.9995 has already worse performance than the model without forgetting.
while an agent's performance is almost 0 when $\gamma$ is 0.999.
This means that even a mere 0.1\% of forgetting in every iteration dramatically affects the performance.

\begin{figure}[t!]
\begin{center}
\includegraphics[width=0.95\linewidth]{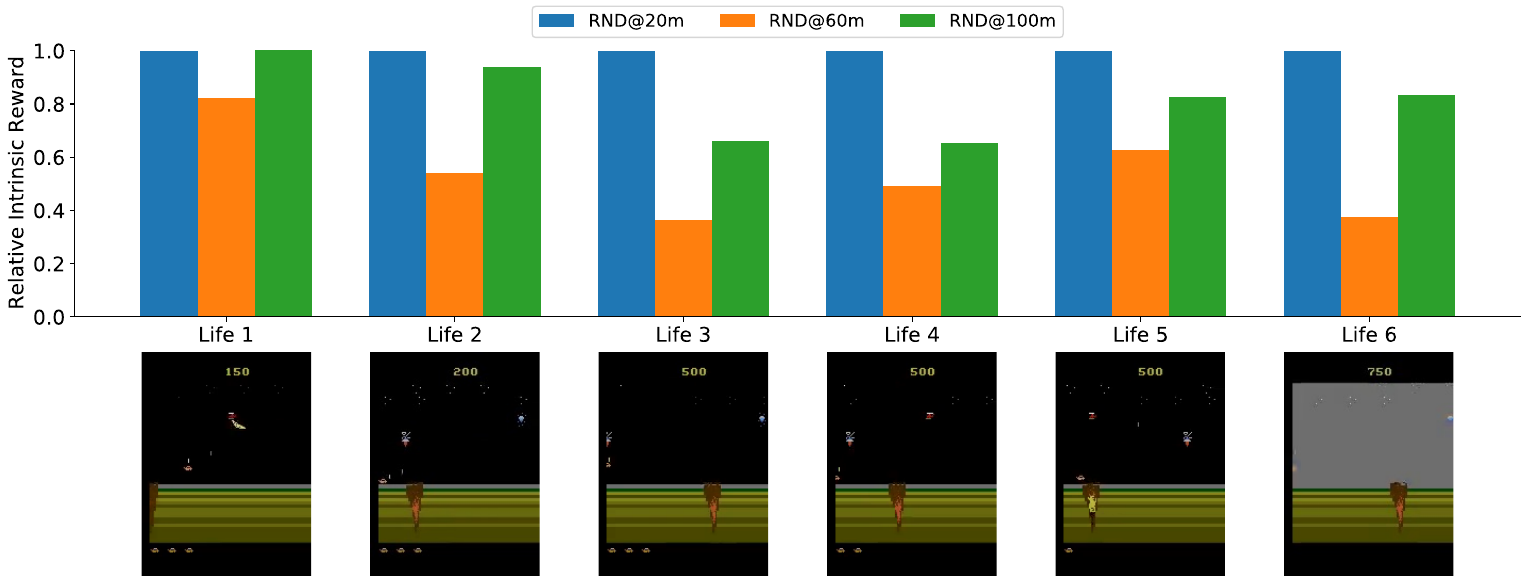}
\end{center}
\caption{
The intrinsic rewards of RND in three different checkpoints during training (training on 20, 60, and 100 million frames, respectively)  when an agent is about to die in a video generated by RND trained on 20 million frames (RND@20m).
Intrinsic reward for each frame is normalized by the original intrinsic rewards generated by RND@20m.
High intrinsic reward denotes that the agent is misguided to die.
In every moment, intrinsic rewards from RND training from 60 million frames (RND@60m) are lower than RND@20m.
However, the rewards are increased again at the end of training (RND@100m).
}
\label{fig:dying}
\end{figure}

\subsection{Catastrophic Forgetting of RND}\label{subsec:forgetting_jamesbond}
we conduct further analysis of RND~\cite{burda2019exploration} in Jamesbond where an agent starts from the same position every episode and moves in one direction without return.
We first generate a trajectory of an episode from a policy after training 20 million frames and then, measure intrinsic rewards of RND at three different checkpoints; after training 20, 60, and 100 million frames, when the agent is dead by missiles, enemies, or obstacles.
As shown in Figure~\ref{fig:dying}, although the intrinsic rewards are decreased in the middle of training, they subsequently increase.
This implies that training decreases the reward but catastrophic forgetting happens between 60 million frames and 100 million frames.
This might be related to the agent more frequently going to the ocean which has a different background and new enemies~(\eg, frogman, mini-submarines) in the later stage of training.
It leads to catastrophic forgetting about the previous scene.


\section{FARCuriosity}\label{sec:method_rl}
\subsection{Overview}\label{subsec:overview}

Intrinsic reward functions~(in the form of curiosity modules)~\cite{burda2019exploration,campero2021learning,pathak2017curiosity,pathak2019self,raileanu2020ride} are used for exploring novel spaces of an environment by generating intrinsic rewards.
The intrinsic reward can serve as the primary reward for training policies in realistic scenarios where external rewards are sparse or absent. 
Since a curiosity module is trained concurrently with the policy, it can suffer from catastrophic forgetting~\cite{french1999catastrophic,robins1995catastrophic} over large and heterogeneous environments where the observation space has high variance such as having multiple stages with different backgrounds as we mentioned in Appendix~\ref{sec:catastrophic}.
Hence, demanding the larger model capacity to memorize what the model has seen for generating intrinsic rewards in these environments.
The catastrophic forgetting of a curiosity module leads to high prediction errors (high intrinsic rewards) for already explored states, only encouraging to visit those states.

We tackle this problem based on the notion of fragmentation and recall, the interaction between short-term memory (STM) inside of an agent (\eg, RAM, VRAM) and long-term memory (LTM) outside of an agent (\ie, external memory; SSD, flash memory).
STM stores the current local curiosity module that generates intrinsic rewards and is updated while an agent explores the environment.
The intrinsic reward (prediction error) is also used as a surprisal which is a criterion of fragmentation.
On the other hand, LTM stores unused curiosity modules.

\begin{figure}[t!]
\begin{center}
\includegraphics[width=0.95\linewidth]{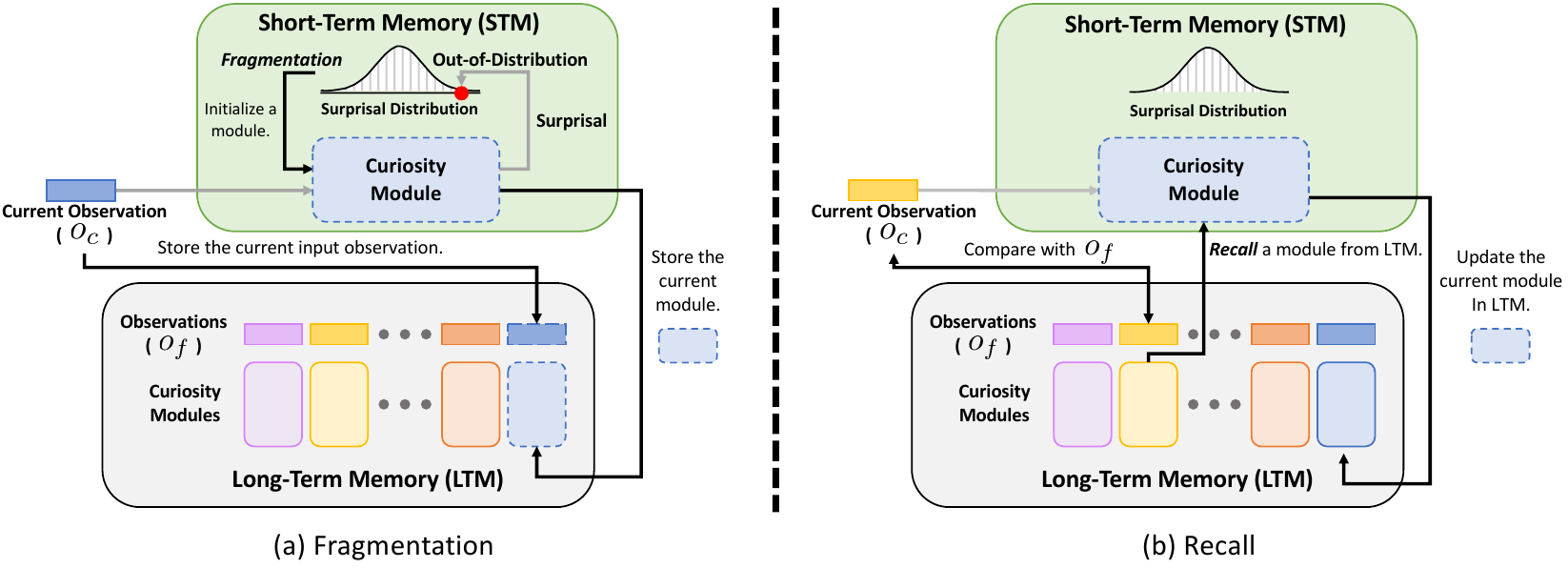}
\end{center}
   \caption{Illustration of the proposed FARCuriosity framework.
   (a) \emph{Fragmentation}. If the current surprisal is out-of-distribution with a high value~(gray arrows), the current curiosity module and its input are stored in LTM and a new module is initialized.
   (b) \emph{Recall}. If the current observation is similar to an observation in LTM, the corresponding curiosity module is recalled and we update the current module in LTM.
}
\vspace{-0.2cm}
\label{fig:architecture_rl}
\end{figure}

When the surprisal~(intrinsic reward) exceeds some threshold, this corresponds to a \emph{fragmentation} event.
At a fragmentation event, the local model is written to long-term memory (LTM) to preserve the local information without forgetting~(Figure~\ref{fig:architecture_rl}a).
The current (abstracted) observation is also stored and used as a key to recall the corresponding model, and the agent initializes an entirely new local model.
On the other hand, if the current observation is sufficiently similar to a stored observation (key) in LTM, the agent \emph{recalls} the corresponding model fragment (local model)~(Figure~\ref{fig:architecture_rl}b).
Hence, the agent can preserve and reuse previously acquired information.
We name this framework fragmentation-and-recall curiosity (\emph{FARCuriosity}).
The goal of {\em FARCuriosity} is to overcome catastrophic forgetting issues in heterogeneous environments to better explore states so that the agent can solve a given task.
We expect this improved exploration to support improved performance on many tasks, above the performance improvements provided by using the existing prediction-based intrinsic-reward approaches that suffer from catastrophic forgetting~\cite{burda2019exploration}

\subsection{Fragmentation}
Fragmentation divides the input space into multiple subspaces and allocates a curiosity module to each subspace.
A fragmentation event occurs  if the ratio of the current surprisal and the running average of the surprisal within the local region is greater than a threshold, $\rho$.
However, $z$-score can be used to incorporate the variance of the surprisal in the subset of the environment. 
When the fragmentation occurs, we store the current curiosity module and its input~(usually the next observation) in LTM and initialize a new module as shown in Figure~\ref{fig:architecture_rl}a.
Consequently, the previous module can preserve the information of the local region without forgetting caused by learning from other spaces.

\subsection{Recall}\label{subsec:recall}
The recall is designed to load the previous curiosity module from LTM when an agent re-enters a local space corresponding to the module.
At every time step, the agent can search LTM to select a curiosity module. 
The curiosity module's input observation (or its encoded feature) is the key used to select a module: 
Given the current input observation, $o_c$, and the stored observation in LTM, of which an element is $o_f$, and a fixed feature extractor $\phi(\cdot)$ (\eg, target network in RND~\cite{burda2019exploration}), we define a recall score $s^\text{rec}_f$ as a cosine similarity between two (encoded) observations.
If $s^\text{rec}_f$ is equal to or greater than a recall threshold $\psi$ for any element $o_f$, we store the current module in LTM and recall the module that corresponds to $o_f$.


\section{Experiments}\label{sec:exp}

\subsection{Experimental Details}
We implement our FARCuriosity method based on RND~\cite{burda2019exploration} and compare it against the vanilla RND and Proximal Policy Optimization~(PPO)~\cite{schulman2017proximal}. Both FARCuriosity and the vanilla RND use PPO for training a policy, as \cite{burda2019exploration} suggested.
We set the fragmentation parameter $\rho$ to $10$, and the recall parameter $\psi$ is $0.99$.
The maximum size of the LTM is $800$, and the least recently used fragment (curiosity module) is substituted by a new fragment if LTM is fully used.
For fragmentation, we also add a cosine similarity constraint: if the cosine similarity between features of the current input and any $o_f$ in LTM is higher than $0.75$, we skip the fragmentation to avoid excessive fragmentation in high dimensional space.
We use a fixed random network in RND as a feature extractor $\phi(\cdot)$ in Section~\ref{subsec:recall} and all curiosity modules share the same fixed network.

\begin{figure}
\centering
\scalebox{0.9}{
     \begin{subfigure}{.2\linewidth}
    \centering
\includegraphics[width=\linewidth]{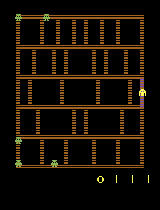}
    \caption{Amidar}\label{fig:amidar}
    \vspace{0.4cm}
\end{subfigure}
\hfill
\hspace{0.3cm}
\begin{subfigure}{.2\linewidth}
    \centering
\includegraphics[width=\linewidth]{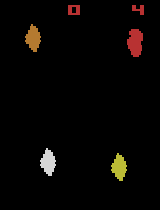}
    \caption{Asteroids}\label{fig:asteroids}
    \vspace{0.4cm}
\end{subfigure}
   \hfill
       \hspace{0.3cm}
\begin{subfigure}{.2\linewidth}
    \centering
\includegraphics[width=\linewidth]{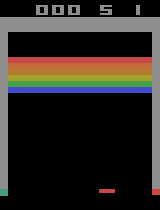}
    \caption{Breakout}\label{fig:breakout}
    \vspace{0.4cm}
\end{subfigure}
\hfill
\hspace{0.3cm}
\begin{subfigure}{.2\linewidth}
    \centering
\includegraphics[width=\linewidth]{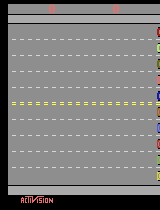}
    \caption{Freeway}\label{fig:freeway}
    \vspace{0.4cm}
\end{subfigure}
}
\scalebox{0.9}{
\begin{subfigure}{.2\linewidth}
    \centering
\includegraphics[width=\linewidth]{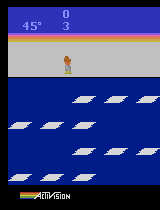}
    \caption{Frostbite}\label{fig:frostbite}
    \vspace{0.4cm}
\end{subfigure}
   \hfill
       \hspace{0.3cm}
\begin{subfigure}{.2\linewidth}
    \centering
\includegraphics[width=\linewidth]{assets/atari/jamesbond.png}
    \caption{Jamesbond}\label{fig:jamesbond}
    \vspace{0.4cm}
\end{subfigure}
   \hfill
       \hspace{0.3cm}
\begin{subfigure}{.2\linewidth}
    \centering
\includegraphics[width=\linewidth]{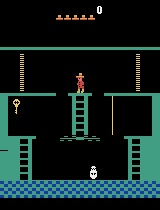}
    \caption{{\tiny Montezuma's Revenge}}\label{fig:monte}
    \vspace{0.4cm}
\end{subfigure}
   \hfill
       \hspace{0.3cm}
\begin{subfigure}{.2\linewidth}
    \centering
\includegraphics[width=\linewidth]{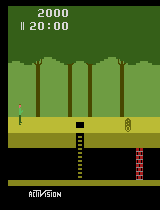}
    \caption{Pitfall}\label{fig:pitfall}
    \vspace{0.4cm}
\end{subfigure}
}
\scalebox{0.9}{
\begin{subfigure}{.2\linewidth}
    \centering
\includegraphics[width=\linewidth]{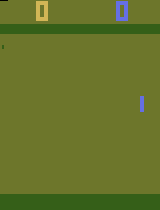}
    \caption{Pong}\label{fig:pong}
    \vspace{0.4cm}
\end{subfigure}
   \hfill
       \hspace{0.3cm}
\begin{subfigure}{.2\linewidth}
    \centering
\includegraphics[width=\linewidth]{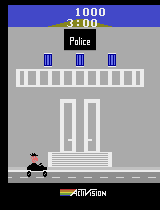}
    \caption{Private Eye}\label{fig:private}
    \vspace{0.4cm}
\end{subfigure}
   \hfill
       \hspace{0.3cm}
\begin{subfigure}{.2\linewidth}
    \centering
\includegraphics[width=\linewidth]{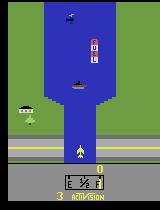}
    \caption{River Raid}\label{fig:riverraid}
    \vspace{0.4cm}
\end{subfigure}
   \hfill
       \hspace{0.3cm}
\begin{subfigure}{.2\linewidth}
    \centering
\includegraphics[width=\linewidth]{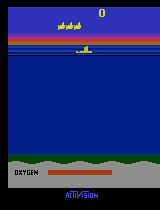}
    \caption{Seaquest}\label{fig:seaquest}
    \vspace{0.4cm}
\end{subfigure}
}
\scalebox{0.9}{
\begin{subfigure}{.2\linewidth}
    \centering
\includegraphics[width=\linewidth]{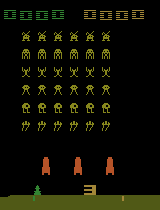}
    \caption{Space Invaders}\label{fig:space}
\end{subfigure}
   \hfill
       \hspace{0.3cm}
\begin{subfigure}{.2\linewidth}
    \centering
\includegraphics[width=\linewidth]{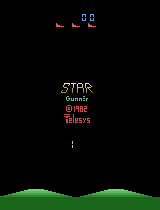}
    \caption{Star Gunner}\label{fig:star_gunner}
\end{subfigure}
   \hfill
       \hspace{0.3cm}
\begin{subfigure}{.2\linewidth}
    \centering
\includegraphics[width=\linewidth]{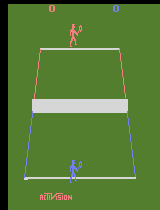}
    \caption{Tennis}\label{fig:tennis}
\end{subfigure}
   \hfill
       \hspace{0.3cm}
\begin{subfigure}{.2\linewidth}
    \centering
\includegraphics[width=\linewidth]{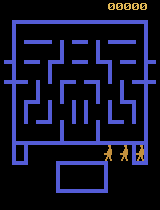}
    \caption{Wizard of Wor}\label{fig:wizard_of_wor}
\end{subfigure}
}
\caption{Various Atari environments used for comparing FARCuriosity and RND.}
\label{fig:atari_env}
\end{figure}

We run Atari experiments for 100 million frames with three different random trials, and we optimize in batches of 128 frames per environment across 32 parallel environments with minibatch size, 4.
We use 16 Atari games: Amidar, Asteroids, Breakout, Freeway, Frostbite, Jamesbond, Montezuma's Revenge, River Raid, Pitfall, Pong, Private Eye, Seaquest, Space Invaders, Star Gunner, Tennis, and Wizard of Wor as shown in Figure~\ref{fig:atari_env}.

\subsection{Comparison with a Baseline}\label{subsec:baseline}

\begin{figure}[t!]
\begin{center}
\centering
\includegraphics[width=0.95\linewidth]{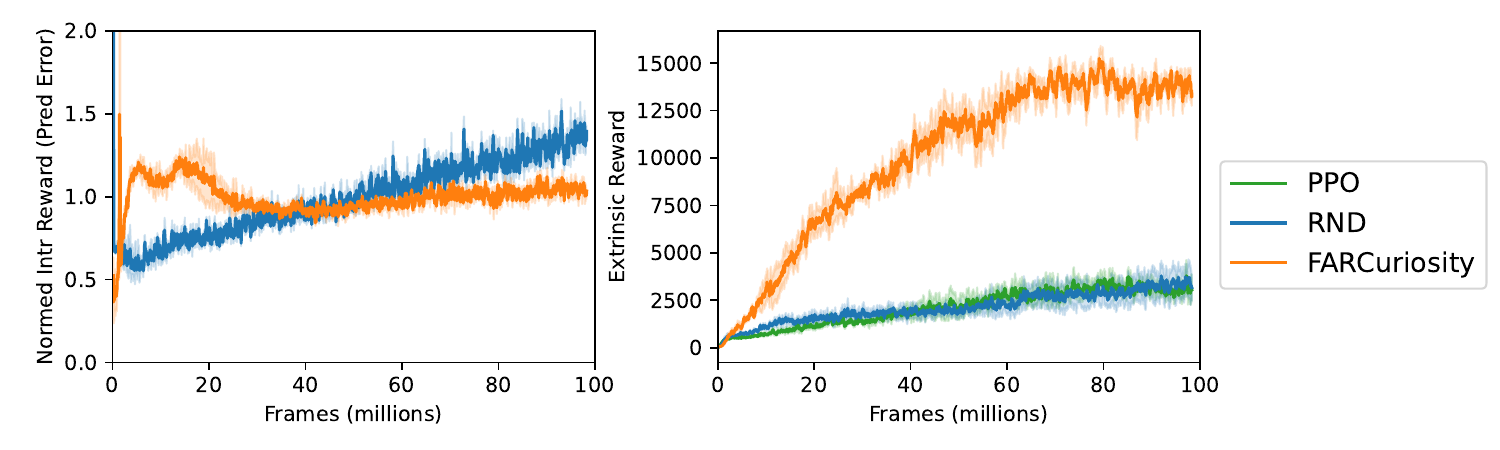}
\end{center}
   \caption{
   (left) Intrinsic of the first observation in every episode (Figure~\ref{fig:env_reset}), and (right) extrinsic reward in Jamesbond during training generated by four different models: PPO, RND, and FARCuriosity.
   Intrinsic reward from RND is increasing over training while FARCuriosity's intrinsic reward is relatively fixed after about 30 million steps.
   On the other hand, FARCuriosity clearly outperforms its counterparts.
}
\label{fig:jamesbond_exp}
\end{figure}

\begin{figure*}[t!]
\begin{center}
\includegraphics[width=0.75\linewidth]{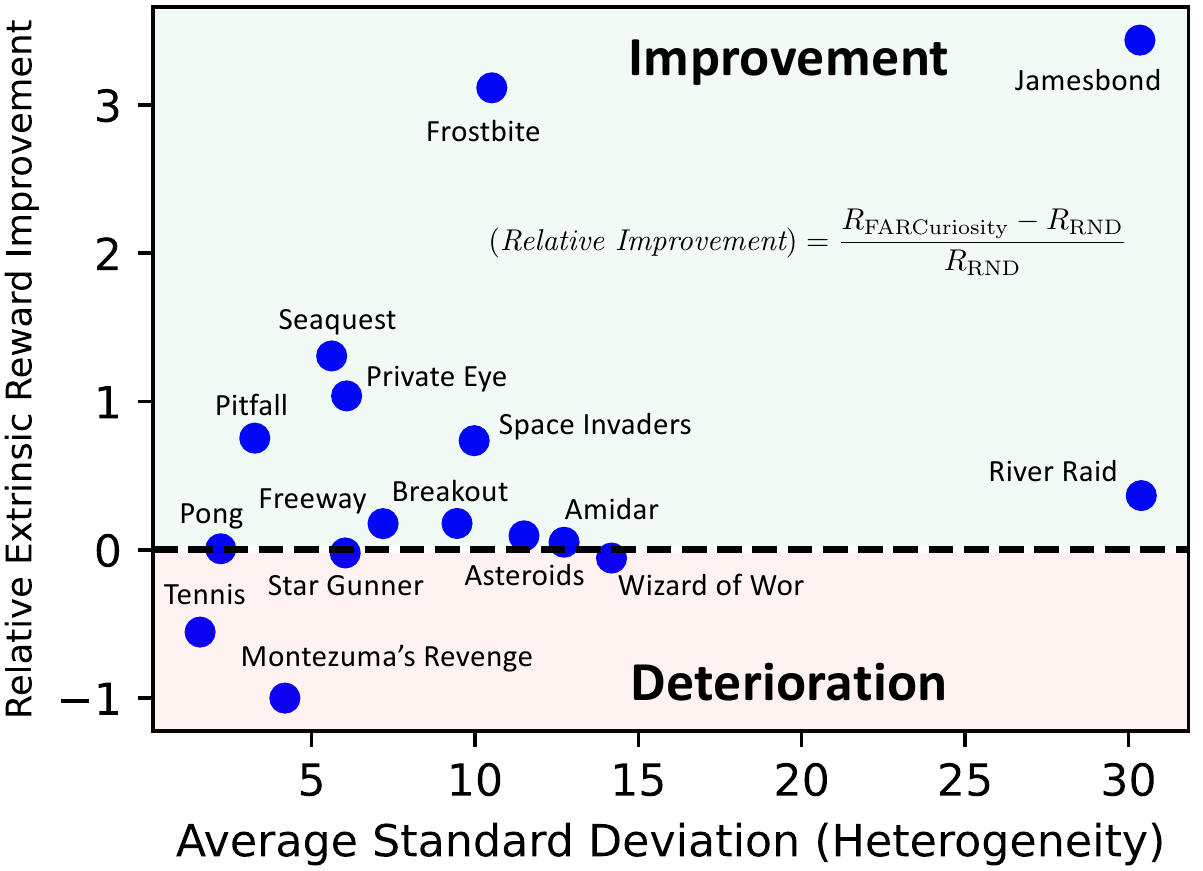}
\end{center}
   \caption{
    The relative performance improvement over RND performance ($R_\text{RND}$) in various Atari games.
    The average standard deviation is measured from each pixel in a trajectory generated by trained RND.
    FARCuriosity is effective in visually high-variance (heterogeneous) games.
}
\label{fig:supp_trend}
\end{figure*}

We measure the intrinsic reward of the first observation in every episode generated by RND during training in Jamesbond in Figure~\ref{fig:jamesbond_exp}.
The intrinsic reward generated by RND increases throughout training, implying that this method suffers from catastrophic forgetting.
On the other hand, the intrinsic reward of the FARCuriosity agent at the observation remains almost the same after a short initial increase, regardless of how much subsequent learning the agent performs elsewhere in the environment.

If the agent only explores a small portion of the environment, fragmentation is not needed although the environment has hundreds of distinctive spaces.
Therefore, we define \textit{heterogeneity} of each Atari environment as the average across all pixels of the standard deviation of each pixel in generated trajectories by a trained policy:
\begin{equation}
(Heterogenity) = \frac{1}{H\times W} \sum_{(h,w) \in (H, W)}\sigma(I_{1 \leq t\leq T,h,w}),
\label{eq:heterogeneity}
\end{equation}
where $\sigma(\cdot)$ denotes standard deviation of its input and  $I\in\mathbb{R}^{T\times H\times W}$ denotes a generated trajectory composed of $T$ frames with size $H \times W$.
We use vanilla PPO~\cite{schulman2017proximal} instead of RND to give a guideline for choosing RND and FARCuriosirty in the environment.

Figure~\ref{fig:supp_trend} measures the relative performance improvement on various Atari games which is defined as below:
\begin{equation}
    (Relative\ Improvement) = \frac{R_\text{FARCuriosity} - R_\text{RND}}{R_\text{RND}},
\end{equation}
where $R_\text{FARCuriosity}$ and $R_\text{RND}$ denote the average extrinsic reward of FARCuriosity and RND, respectively.
Since both $R_\text{FARCuriosity}$ and $R_\text{RND}$ are negative in Tennis and Pitfall, we multiply the improvement by -1 here.
The negative improvement denotes that performance is worsened by FARCuriosity.
FARCuriosity generally improves performance in heterogeneous environments while achieving similar or even worse performance in homogeneous games.
FARCuriosity achieves worse performance in Montezuma's Revenge which might be due to its low heterogeneity since PPO could not explore many rooms.
In the least heterogeneous environment, Tennis where only players and a ball are moving with a fixed background, FARCuriosity deteriorates compared to RND.
We analyze the relationship between the number of fragments and the performance in Appendix~\ref{sec:supp_n_frag} and present the average game score for each model in each environment in Table~\ref{tab:results} and Figure~\ref{fig:results_rl} in the appendix.

\subsection{Effect of the Number of Fragments}\label{sec:supp_n_frag}
\begin{figure}[t!]
\begin{center}
\begin{subfigure}{0.47\textwidth}
\centering
\includegraphics[width=\linewidth]{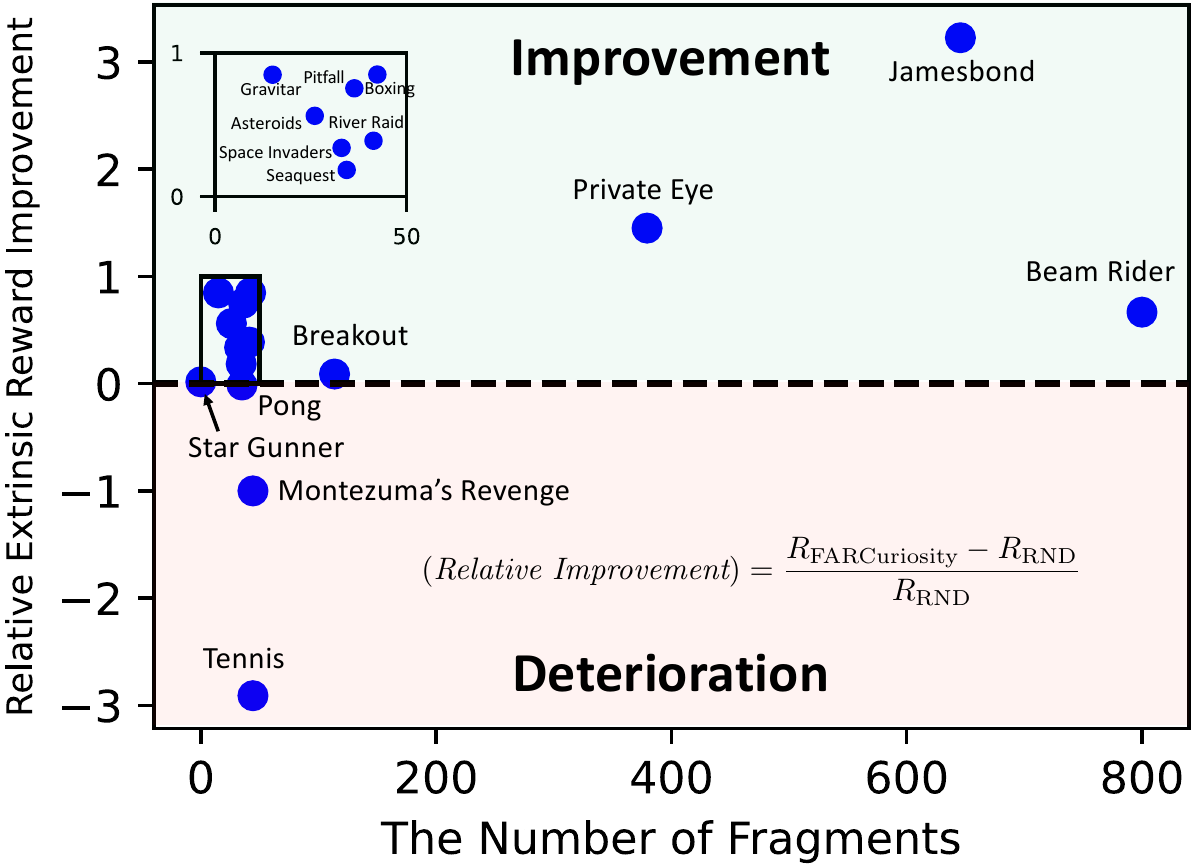}
\caption{}
\label{fig:jamesbond_intr}
\end{subfigure}
\begin{subfigure}{0.47\textwidth}
\centering
\includegraphics[width=\linewidth]{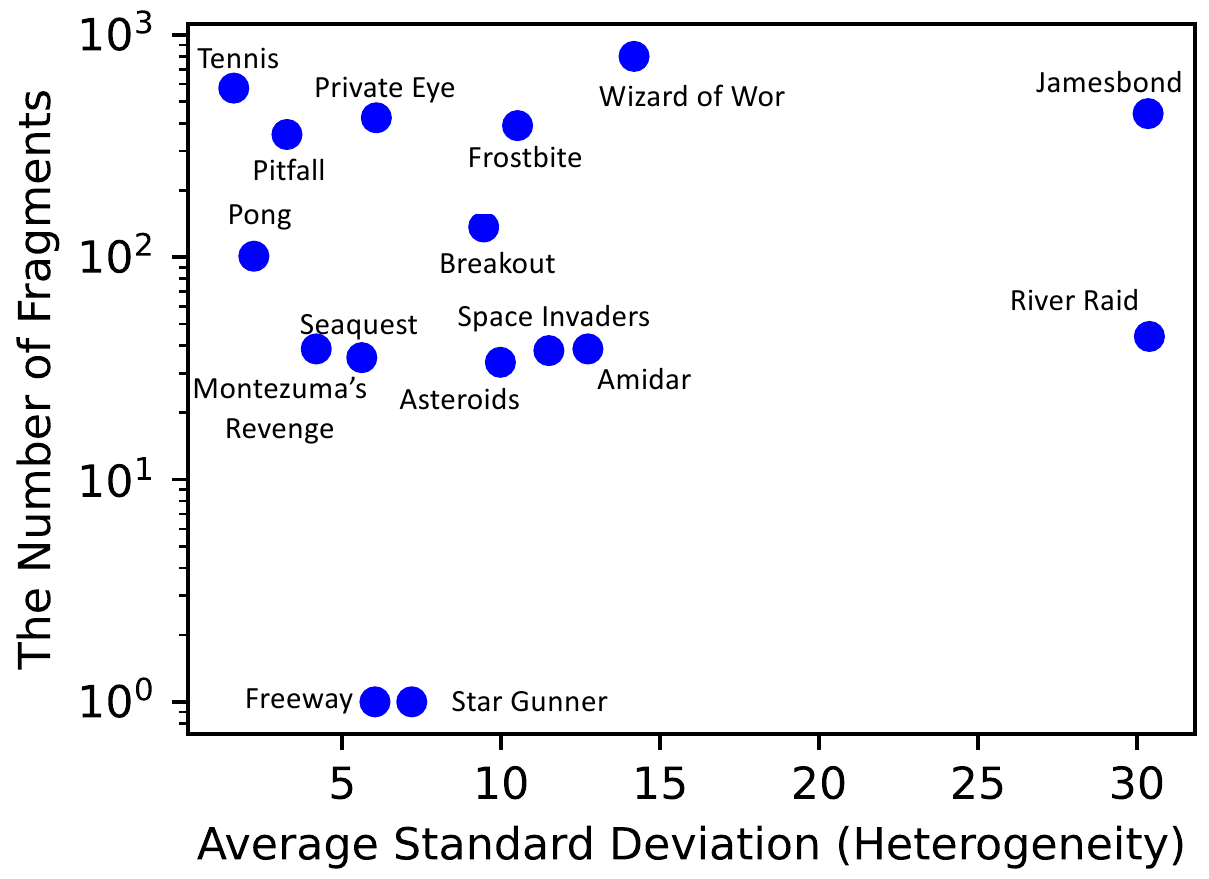}
\caption{}
\label{fig:hetero_nfrag}
\end{subfigure}
\end{center}
\vspace{-0.15cm}
   \caption{
    (a) The relative performance improvement over RND performance ($R_\text{RND}$) in various Atari games.
    (b)     The number of fragments is not related to the heterogeneity of the game.
}
\label{fig:frag_trend}
\end{figure}

Figure~\ref{fig:frag_trend} shows the relation between the relative performance improvement and the average number of fragments on multiple Atari games.
In most games, the number of fragments is up to 200, but it reaches the maximum capacity of LTM in Wizard of Wor.
There is no clear global tendency between performance improvement and the number of fragments.
On the other hand, the number of fragments is not related to the heterogeneity of games as shown in Figure~\ref{fig:hetero_nfrag}. 
This is because we use a dynamic threshold using running statistics of surprisal to avoid over-fragmentation in highly heterogeneous environments.
However, it triggers excessive fragmentation in low heterogeneous environments as a trade-off.

\section{Discussion}\label{sec:conclusion}

We shed light on the problem of catastrophic forgetting when using prediction-based intrinsic rewards in reinforcement learning.
Catastrophic forgetting happens not only in reset-free environments but also in resetting environments as an agent explores more observation spaces and lives longer.
The forgetting leads to a collapse of the assumption in the intrinsic reward function: the function is a monotonically decreasing function on the same observation over time.
Consequently, it can potentially yield a problem: the agent is prone to visit already-visited states.
We also demonstrate the hazard of catastrophic forgetting using count-based exploration with exponential decay in a minigrid environment.

To alleviate the catastrophic forgetting problem, we propose a simple method inspired by how natural agents explore space.
Our FARCuriosity \emph{fragments} the exploration space (\ie, observation space) based on the surprisal, which is generated by a curiosity module~\cite{burda2019exploration} in short-term memory (STM), in real-time.
During fragmentation, an agent stores the current curiosity module in long-term memory~(LTM) and the stored module \emph{recalled} when the agent returns to the state where the fragmentation happened so that the agent can reuse the local information.
Consequently, FARCuriosity achieves better performance in heterogeneous environments while it has drawbacks in homogeneous environments.
However, we did not assess fragmentation quality and leave it for future study.
Our paper aims to highlight the catastrophic forgetting issue in prediction-based intrinsic reward function, which is potentially hazardous to RL agents by misguiding behavior.

\section*{Acknowledgement}
Ila Fiete is supported by the Office of Naval Research, the Howard Hughes Medical Institute~(HHMI), and NIH (NIMH-MH129046).

\bibliographystyle{ieee_fullname}
\bibliography{egbib}

\clearpage

\section*{\large{Appendix}}
\setcounter{section}{0}
\renewcommand\thesection{\Alph{section}}


\begin{figure}[h!]
\begin{center}
\begin{subfigure}{0.25\textwidth}
\centering
\includegraphics[width=\linewidth]{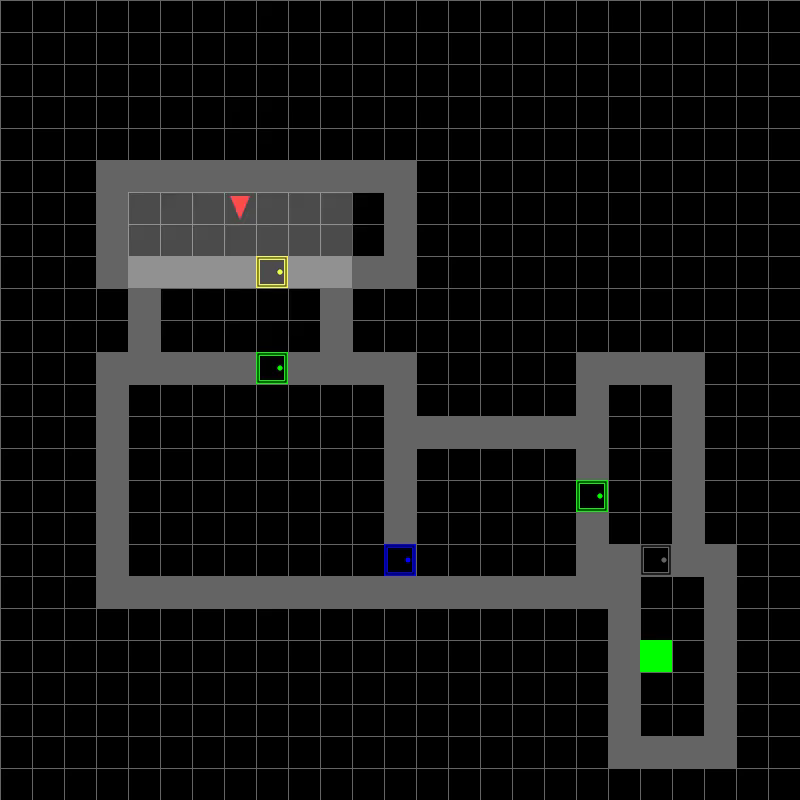}
\end{subfigure}
\hspace{1cm}
\begin{subfigure}{0.25\textwidth}
\centering
\includegraphics[width=\linewidth]{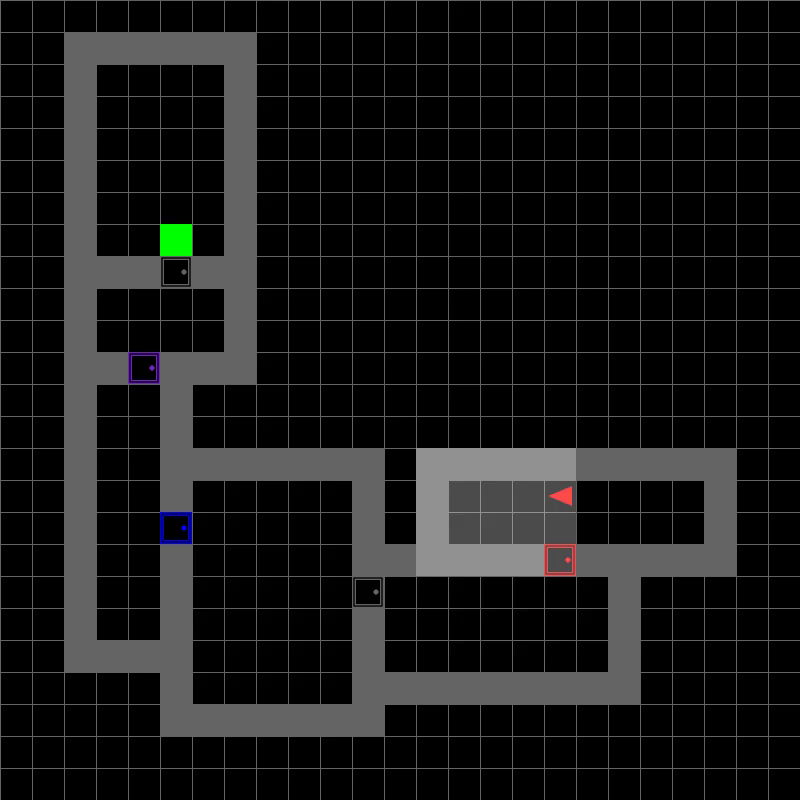}
\end{subfigure}
\hspace{1cm}
\begin{subfigure}{0.25\textwidth}
\centering
\includegraphics[width=\linewidth]{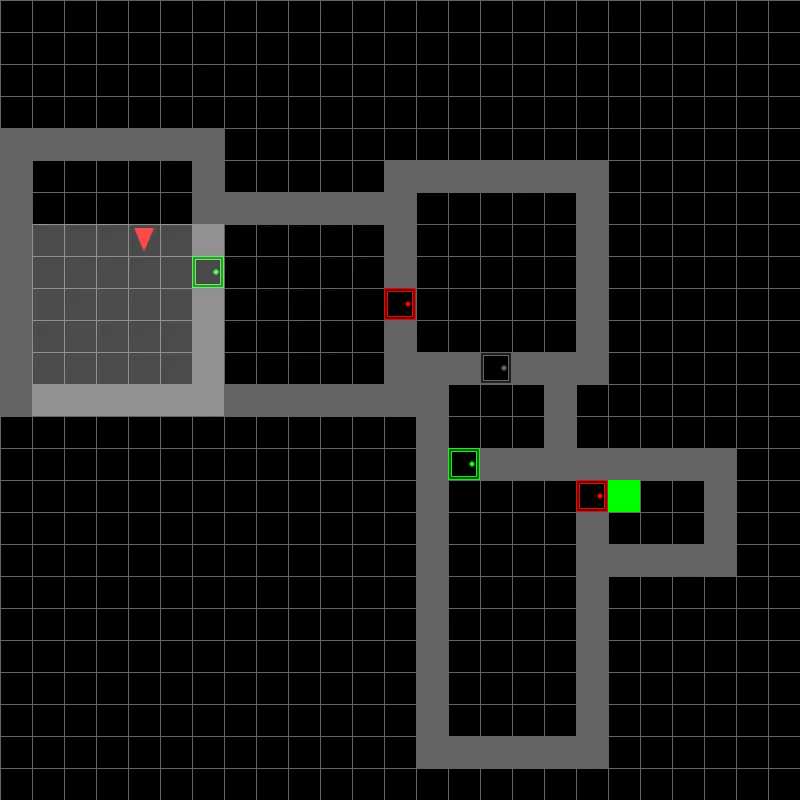}
\end{subfigure}
\end{center}
   \caption{
  Examples of \textsf{MultiRoom-N6} in MiniGrid benchmark.
  The red triangle is an agent and the green rectangular is the goal location.
  A rectangular with a dot is a door the agent needs to open.
   }
\label{fig:minigrid_env}
\end{figure}

\section{Experimental Details}\label{subsec:exp_details}
Our model is implemented on PyTorch and the experiments are conducted on Intel(R) Xeon(R) CPU E5-2650 v4 @ 2.20GHz and a single NVIDIA Tesla V100 GPU for reinforcement learning experiments.

\subsection{Main Experiments}
The learning rate is 0.0001, the reward discount factor is 0.99 and the number of epochs is 4.
For other parameters, we use the same values mentioned in PPO and RND: we set the GAE parameter $\lambda$ as 0.95, value loss coefficient as 1.0, entropy loss coefficient as 0.001, and clip ratio~($\epsilon$ in Eq. (7) in \cite{schulman2017proximal}) as 0.1.
\subsection{Toy Grid World}
\label{subsec:supp_toy}
We conducted simple experiments on a toy grid environment to understand why catastrophic forgetting happens even in an environment that is reset multiple times during training.
The size of the grid environment is $10 \times 10$ and an observation of each location is a 32-dimensional vector sampled from a uniform distribution, $\mathcal{U}(0, 1)$.
In every episode, the agent starts from the center of the environment $(5,5)$ and randomly explores the environment except the start position.
The episode length is 200 in a fixed length setting (Figure~\ref{fig:grid_example_fixed}) and $10n$ at $n$-th episode in the gradually increasing setting~(Figure~\ref{fig:grid_example_increased}).
Both fixed and training networks in RND are implemented as two fully connected layers with Rectified Linear Unit (ReLU).

We use Adam~\cite{kingma2015adam} with a learning rate of 0.001 and RND is sequentially trained on the current observation (the number of the parallel environments is 1 and the batch size is 1) for one million steps.
We used three different random seeds and calculate means and standard errors in Figure~\ref{fig:grid_example}.

\subsection{Count-Based Exploration in MiniGrid benchmark}\label{subsec:supp_minigrid}

MiniGrid~\cite{gym_minigrid} is a procedurally-generated grid world environment with various tasks.
The observation is $7\times 7$ egocentric view with three channels (cell types, cell colors, and door states).
The number of actions is seven: 
turn left, turn right, move forward,
pick up an object, drop an object, toggle (open/close a door), and done.
\textsf{MultiRoom-N6} is composed of six rooms.
An agent needs to navigate from the first room to the goal location in the last room passing other rooms as shown in Figure~\ref{fig:minigrid_env}.
The agent needs to open (toggle) doors to enter the next room.
The episode length $t_{\max}$ is 120 and the extrinsic reward is defined as $1- 0.9t/t_{\max}$ at time $t$.

We use MD5 hash for counting the number of visits of each observation and keep updating it across multiple episodes during the training.
Since the observation is  $7\times 7$ egocentric view following previous works~\cite{campero2021learning,raileanu2020ride}, multiple locations can share the same observation and the same intrinsic reward.
We set the exponential decay factor, $\gamma$ as 0.9990, 0.9995, and 1.0000, and run each model with three different trials for 10 million steps.
Other hyperparameters are the same as the main experiments.

\section{Sensitivity Analysis}
We conducted sensitivity analysis for fragmentation and recall thresholds for 50 million frames in Jamesbond as shown in Table~\ref{tab:sensitivity}.
While changing one threshold, we fix another threshold.
A small fragmentation threshold reduces the performance due to excessive fragmentations while FARCuriosity is robust to the change of recall threshold.
However, the proposed method still achieves better performance than baselines.

\begin{table}
\centering
\caption{Sensitivity analysis about fragmentation threshold, $\rho$ and recall threshold, $\psi$ in Jamesbond.}
\vspace{0.2cm}
\begin{tabular}{c|c}
\toprule
Model & Average Score \\ \hline
PPO~\cite{schulman2017proximal} & 2022.9 \\
RND~\cite{burda2019exploration} & 1925.6 \\  \midrule
FARCuriosity ($\rho = 5$) & 5339.8 \\
FARCuriosity ($\rho = 10$, ours) & \textbf{11917.5} \\
FARCuriosity ($\rho = 15$) & 10407.8 \\ \midrule
FARCuriosity ($\psi = 0.95$) & \textbf{12719.5} \\
FARCuriosity ($\psi = 0.975$) & 11917.5 \\
FARCuriosity ($\psi = 0.99$, ours) & 10407.8 \\
\bottomrule
\end{tabular}
\label{tab:sensitivity}
\end{table}

\section{Individual Performance in Atari Games}\label{sec:supp_atari_each}

\begin{figure*}[t]
\begin{center}
\includegraphics[width=0.95\linewidth]{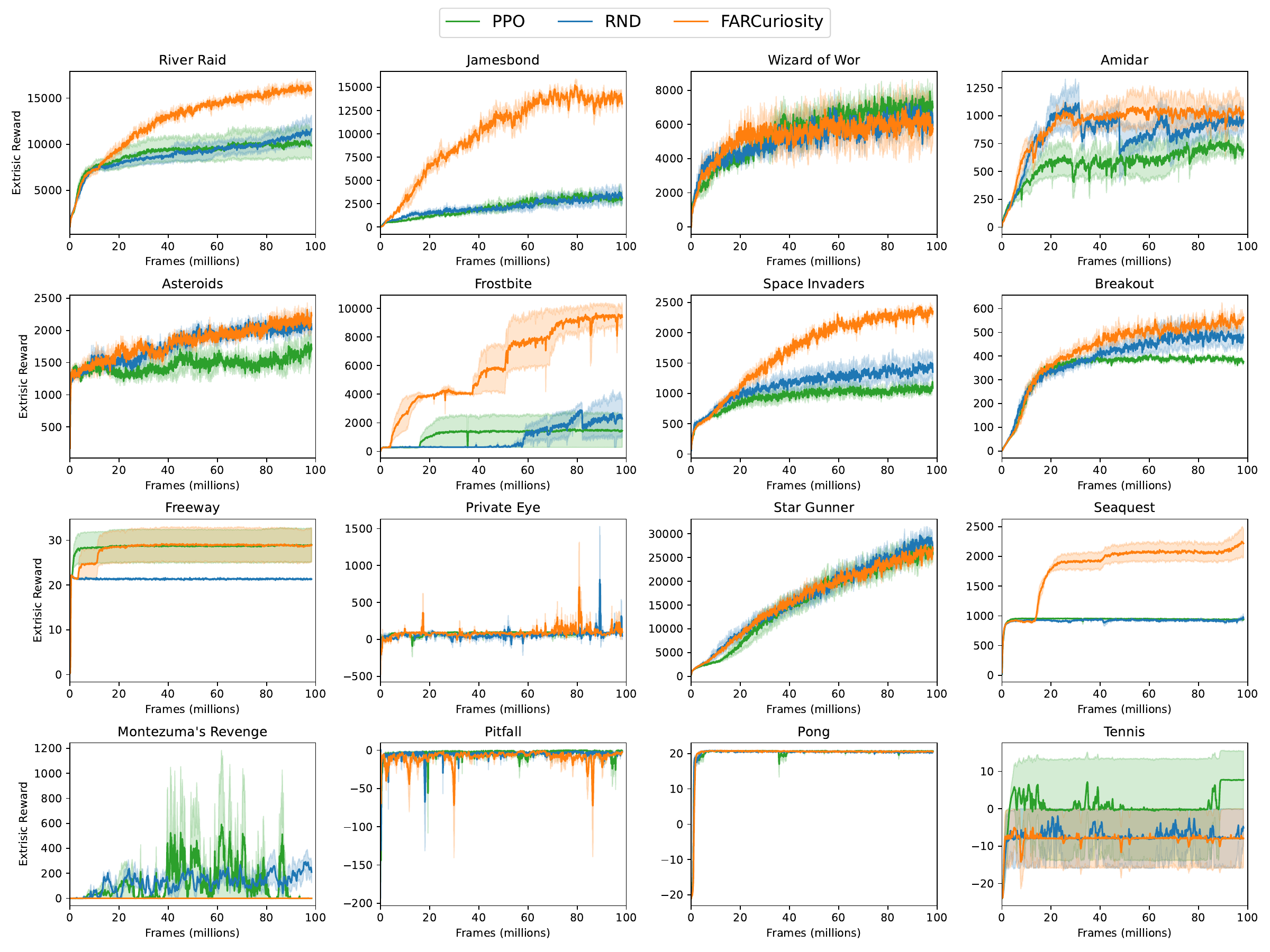}
\end{center}
   \caption{
   Mean extrinsic reward of RNN-based policies: PPO, RND, and FARCuriosity with extrinsic rewards on sixteen Atari games.
The environments are sorted by the heterogeneity measured in Section~\ref{sec:exp}: from highest to lowest.
   }
\label{fig:results_rl}
\end{figure*}

Figure~\ref{fig:results_rl} presents the mean cumulative extrinsic rewards (also known as extrinsic return) and its standard errors at varying numbers of training data (frames) over three different random trials in sixteen games including three hard Atari games: Montezuma's Revenge, Pitfall, and Private Eye.
No fragmentation occurs in Freeway and Star Gunner, which means that the difference between RND and FARCuriosity in those environments is because of the randomness of reinforcement learning and the environments.
FARCuriosity outperforms RND in eight environments.
These environments mainly have high heterogeneity except Seaquest.
Especially, Jamesbond and River Raid have much more heterogeneity compared to other environments, and FARCuriosity achieves  clearly better performance.
In Breakout, the performance gap is much smaller.
This is because the observations in those environments are more similar, and once the agent hits all bricks, the same  bricks are revived.
Hence, RND does not suffer from catastrophic forgetting and achieves similar results to FARCuriosity.
FARCuriosity shows similar performance in Pitfall, Pong, Star Gunner, and Tennis.
We believe this is because the observation space is less visually diverse in those games.
In Pong, for example, the only visually dynamic features are the agent, the opponent, and the ball, all of which move over a fixed monotone background.
On the other hand, FARCuriosity shows worse performance compared to RND in Montezuma's Revenge.
Montezuma's Revenge is one of the most homogeneous games according to our measurement since an agent is difficult to explore many rooms.
We hypothesize that this homogeneity can lead to wrong fragmentation, which degrades the performance.
There are about 600 fragments in Tennis but it does not change the performance since the environment is easy while Montezuma's Revenge is not.
We also measure Interquartile Mean~(IQM)~\cite{agarwal2021deep} as aggregated human-normalized scores across environments.
FARCuriosity outperforms its baselines.

\begin{table}
\centering
\caption{The average score of each method in Atari games with their heterogeneity. The numbers in parentheses are human-normalized scores where human performance is acquired by Badia~\etal~\cite{badia2020agent57}.}
\vspace{0.3cm}
\begin{tabular}{c|c|ccc}
\toprule
Game & Heterogeneity & PPO & RND & FARCuriosity \\ \hline
River Raid& 30.4	&9860.9~(54)& 	11659.3~(65.4)	& \textbf{15923.6~(92.4)}\\
Jamesbond&30.4 	&3060.7~(1107.3) &	2946.2~(1065.4)	& \textbf{13073.0~(4764.1)}\\
Wizard of Wor& 14.2	&\textbf{7392.0~(162.9)}	& 6123.0~(132.6) &5782.3~(124.5)\\
Amidar& 12.7	& 648.0~(37.5)	 &963.8~(55.9)	& \textbf{1012.9~(58.8)}\\
Asteroids&  11.5	&1724.6~(2.2)	& 2028.5~(2.8)& 	\textbf{2219.7~(3.2)}\\
Frostbite& 10.5	&1444.7~(32.3)	& 2282.0~(51.9)& 	\textbf{9392.1~(218.5)}\\
Space Invaders& 10.0	&1088.7~(61.9) & 	1352.6~(79.2)	&\textbf{2347.9~(144.7)}\\
Breakout&  9.5  & 377.3~(1304.2)& 	478.5~(1655.6)& 	\textbf{563.5~(1950.7)} \\
Freeway	&   7.2 &\textbf{29.0~(98.0)}&  21.3~(72.0)& 	25.1~(84.8)\\
Private Eye&6.1 	&97.0~(0.1)	& 47.6~(0.0)	& \textbf{97.0~(0.1)}\\
Star Gunner& 6.0	&25982.7~(264.1)& 	\textbf{27063.7~(275.4)}	&26491.3~(269.4)\\
Seaquest& 5.6	    &936.7~(2.1) & 	961.6~(2.1)& 	\textbf{2218.8~(5.1)}\\
Montezuma Revenge& 4.2	&0.0~(0.0)	& \textbf{210.3~(4.4)}& 	0.0~(0.0)\\
Pitfall& 3.3	&\textbf{-0.7~(3.4)}	& -4.1~(3.4)	& -1.0~(3.4)\\
Pong& 2.2	&\textbf{20.8~(117.6)}	& 20.5~(116.7)	& 20.6~(117)\\
Tennis& 1.6	&\textbf{7.8~(203.9)}	& -5.1~(120.6)& 	-8.0~(101.9)\\\midrule
\multicolumn{2}{c|}{IQM~\cite{agarwal2021deep}} & 0.69 & 0.68 & \textbf{0.91} \\
\bottomrule
\end{tabular}
\label{tab:results}
\end{table}

\end{document}